\documentclass[PRA,nofootinbib,superscriptaddress,a4paper, twocolumn,floatfix,longbibliography]{revtex4-2}

\usepackage[english]{babel}
\usepackage[utf8]{inputenc}
\usepackage[colorinlistoftodos, color=green!40, prependcaption]{todonotes}
\usepackage{amsthm}
\usepackage{mathtools}
\usepackage{physics}
\usepackage{listings}
\usepackage{xcolor}
\usepackage{graphicx}
\usepackage[left=23mm,right=13mm,top=35mm,columnsep=15pt]{geometry} 
\usepackage{adjustbox}
\usepackage{placeins}
\usepackage[T1]{fontenc}  
\usepackage{booktabs}
\usepackage{amssymb}
\usepackage{bm}
\usepackage[ruled,vlined]{algorithm2e}
\usepackage[caption=false]{subfig}

\usepackage{float}

\usepackage{tabularx}
\usepackage{multirow}
\usepackage{booktabs}

\usepackage[colorlinks=true, linkcolor=blue, citecolor=gray]{hyperref}

\makeatletter
\newcommand*{\rom}[1]{\expandafter\@slowromancap\romannumeral #1@}
\makeatother

\begin{document}

\title{\textbf{Discovering emergent connections in quantum physics research \\ via dynamic word embeddings}}

\author{Felix Frohnert}
\email[E-mail:]{f.frohnert@liacs.leidenuniv.nl}
\affiliation{$\langle a Q a^L \rangle$ Applied Quantum Algorithms, Universiteit Leiden, The Netherlands}

\author{Xuemei Gu}
\affiliation{Max Planck Institute for the Science of Light, Erlangen, Germany}

\author{Mario Krenn}
\affiliation{Max Planck Institute for the Science of Light, Erlangen, Germany}

\author{Evert van Nieuwenburg}
\affiliation{$\langle a Q a^L \rangle$ Applied Quantum Algorithms, Universiteit Leiden, The Netherlands}


\begin{abstract}
As the field of quantum physics evolves, researchers naturally form subgroups focusing on specialized problems. 
While this encourages in-depth exploration, it can limit the exchange of ideas across structurally similar problems in different subfields. 
To encourage cross-talk among these different specialized areas, data-driven approaches using machine learning have recently shown promise to uncover meaningful connections between research concepts, promoting cross-disciplinary innovation.
Current state-of-the-art approaches represent concepts using knowledge graphs and frame the task as a link prediction problem, where connections between concepts are explicitly modeled.
In this work, we introduce a novel approach based on dynamic word embeddings for concept combination prediction. 
Unlike knowledge graphs, our method captures implicit relationships between concepts, can be learned in a fully unsupervised manner, and encodes a broader spectrum of information. 
We demonstrate that this representation enables accurate predictions about the co-occurrence of concepts within research abstracts over time.
To validate the effectiveness of our approach, we provide a comprehensive benchmark against existing methods and offer insights into the interpretability of these embeddings, particularly in the context of quantum physics research. 
Our findings suggest that this representation offers a more flexible and informative way of modeling conceptual relationships in scientific literature.
\end{abstract}

\maketitle

\section{Introduction}
The corpus of scientific literature is expanding at an ever-increasing rate. 
In particular, the field of quantum physics is witnessing a steady growth in published papers each year, growing at an increasing pace.\footnote{Judged by the yearly number of \emph{quant-ph} papers submitted to the arXiv, visualized in Appx.~\ref{appx:papers_per_year}.}
Thus, it is becoming increasingly challenging for individual researchers to gain a comprehensive understanding of the diverse domains within quantum physics. 
Naturally, this leads to the formation of various specialized subgroups within the community~\cite{jacobsSpecializationSynthesisProliferation2014}. 
While specialization enables researchers to focus on specific areas, it also creates isolated knowledge silos where valuable insights may not be shared across different fields. This compartmentalization can result in parallel research trajectories, where solutions to structurally similar problems could remain confined within particular subfields. 
Overcoming this isolation can help researchers from leveraging advancements in adjacent fields that could accelerate progress in their own field~\cite{evans2011metaknowledge,fortunato2018science, wang2021science}. 

Recent efforts to bridge the knowledge gaps between specialized subfields have led some researchers to leverage large language models (LLMs) trained on vast amounts of scientific literature. 
By feeding millions of papers into these models, the aim is to generate or rank novel research ideas directly~\cite{wang2023scimon, baek2024researchagent, guGenerationHumanexpertEvaluation2024, si2024can}. 
While this approach holds significant promise, it poses substantial challenges in evaluation -- it inherently requires human expertise to assess the quality and feasibility of the generated ideas.

An alternative strategy involves forecasting the future trajectory of scientific research by predicting what scientists might work on next or identifying areas poised for high impact~\cite{clauset2017data}. 
A common way to solve meta-scientific forecast tasks is to model the scientific progress as an evolving knowledge graph constructed from millions of publications, closely aligning with link prediction problems in network theory~\cite{liben2003link}. 
Pioneered in the context of biochemistry, it has been demonstrated that knowledge graphs can be used to find more efficient research strategies to accelerate collective scientific discovery~\cite{rzhetsky2015choosing}. 
This idea has been expanded to the forecast of research directions in quantum physics~\cite{krennPredictingResearchTrends2020a} and artificial intelligence~\cite{krennForecastingFutureArtificial2023a}, the prediction of future impact of new scientific connections~\cite{guForecastingHighimpactResearch2024} and to the predictions of interesting research directions which human researchers might not discover~\cite{guGenerationHumanexpertEvaluation2024}.
However, representing scientific literature as a knowledge graph reduces the information down to binary connections between concepts, which may not fully capture the intricate dynamics of scientific advancement.

One way to address this issue is by using unsupervised word embeddings derived \emph{directly} from scientific literature, analyzing how these embeddings evolve to predict future research trends and dynamics.
\begin{figure*}[!t]
    \includegraphics[width=\textwidth]{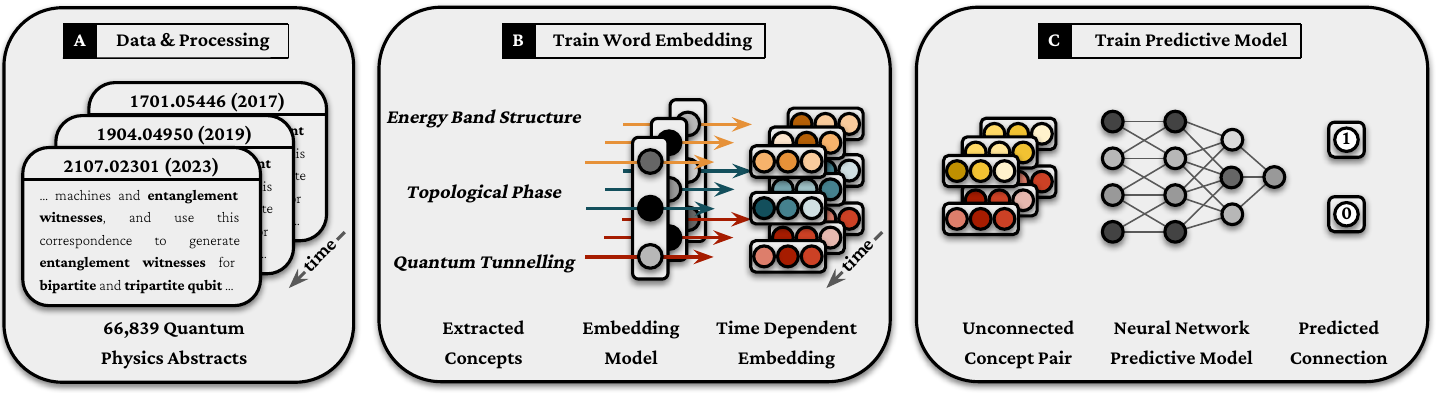} 
    \caption{\textbf{Overview:} 
    (a) We analyze a dataset of $66,839$ papers with the \textit{quant-ph} identifier on arXiv, spanning from $1994$ to $2023$. 
    From these papers, we extract $10,235$ quantum physics-related concepts using RAKE and other NLP tools.
    (b) Using the abstracts of these papers, we train an embedding model to capture the evolving relationships between these concepts in vector representations over time. 
    In the visualization, gray dots indicate changes in the embedding model’s weights over the years, while the hues of orange, cyan, and red represent the dynamics of word embeddings' parameters as they change with time.
    (c) The task involves training a machine learning model to predict which currently unconnected concepts (those not yet studied together) are likely to co-occur in the near future, based on the learned embeddings.
    }
    \label{fig:overview}
\end{figure*}
This approach fundamentally differs from knowledge graphs by capturing information from scientific literature without human bias, representing words solely based on contextual similarities and mapping semantically related terms to closely related points in the embedding.
Prominent examples involve the automated encoded of scientific literature in material science. 
Here, when the names and properties of materials are encoded, the word embedding can be used to find new materials with specific properties~\cite{tshitoyanUnsupervisedWordEmbeddings2019,KimInorganic2020,ShettyAutomated2021}. 
Another work demonstrated how automated embeddings can relate the surprise and impact of research ideas~\cite{shi2023surprising, sourati2023accelerating}. 

In our work, we demonstrate how an unsupervised dynamic word embedding can efficiently predict future research directions in quantum physics. 
Without relying on any hand-crafted features, a machine learning model that uses only the coordinates of the embedded latent space can predict the likelihood that previously unstudied research combinations will be explored in the future. 
Our method surpasses other approaches that do not rely on human-crafted features, showcasing a pathway for fully end-to-end prediction tasks within the science of science.

The rest of the paper is structured as follows: 
Sec.~\ref{sec:method} discusses the methodology, including embedding generation process and the machine learning pipeline.
Sec.~\ref{sec:result} presents the benchmarking results of the trained ML models and interpretation of the learned embedding. 
Finally, Sec.~\ref{sec:discussion} discusses the findings and provides an outlook for future research.

\begin{figure*}[ht!]
    \centering
    \begin{tabular}{cc}
        \includegraphics[width=0.45\linewidth]{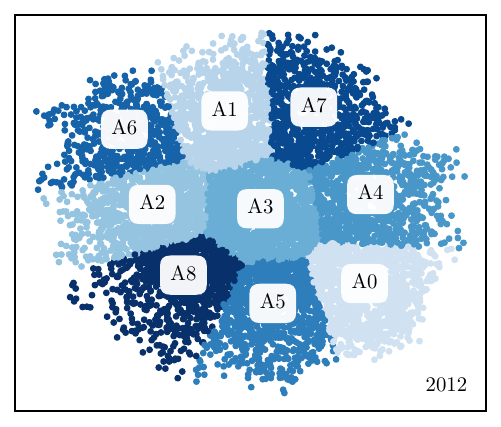} &
        \includegraphics[width=0.45\linewidth]{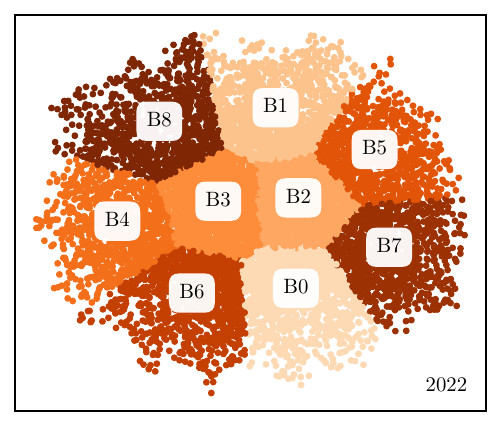} \\
        \begin{tabularx}{0.424\textwidth} { 
            | >{\centering\arraybackslash}p{0.025\linewidth} 
            | >{\raggedright\arraybackslash}X 
            | >{\raggedright\arraybackslash}X 
            | >{\raggedright\arraybackslash}X | } 
            \hline
            A0 & quantum memory & quantum processor &  quantum computation \\
            \hline
            A1 & erasure \ \ \ channel & butterfly network & quantum fingerprinting \\
            \hline
            A2 & gallium arsenide & nitrogen \ \ \ atom &  quantum dot emission \\
            \hline
            A3 & gaussian wavepacket & quantum propagator &  coherent \ \ \ state \\
            \hline
            A4 & phase \ \ \ \ \ \ \ \ \ noise & laser \ \ \ linewidth &  cavity resonance \\
            \hline
            A5 & compton scattering & kapitza dirac effect &  electron interferometer \\
            \hline
            A6 & ornstein uhlenbeck & white gaussian noise &  stochastic dynamic \\
            \hline
            A7 & clifford \ \ \ \ \ \ group &  quantum compiler &  anyonic \ \ \ model \\
            \hline
            A8 & wavelength multiplexing & laser \ \ \ pumping & quantum repeater \\
            \hline
        \end{tabularx} &
        \begin{tabularx}{0.424\textwidth} { 
            | >{\centering\arraybackslash}p{0.025\linewidth} 
            | >{\raggedright\arraybackslash}X 
            | >{\raggedright\arraybackslash}X 
            | >{\raggedright\arraybackslash}X | } 
            \hline
            B0 & electric \ \ \ \ \ \ \ \ \ \  field & charged particle & electron \ \ \ \ \  wave \\
            \hline
            B1 & ground state wave function & exact diagonalization &hartree fock theory \\
            \hline
            B2 & quantum computer & quantum speedup &  quantum advantage \\
            \hline
            B3 & silicon quantum dot & electron spin resonance & metal oxide semiconductor  \\
            \hline
            B4 & photonic integration & satellite communication  &  network capacity \\
            \hline
            B5 & dressed \ \ \ \ \  state & rabi \ \ \ \ \ oscillation & vacuum splitting \\
            \hline
            B6 & quantum chaotic & prethermal state  &  ergodic \ \ \ \ \  system \\
            \hline
            B7 & quantum \ \ \ \ \  logic & boolean \ \ \ \ \ algebra &  logical connective \\
            \hline
            B8 & laser interferometer & gravitational wave &  squeezed quadrature \\
            \hline
        \end{tabularx} \\
        \end{tabular}    
    \caption{
    \textbf{Clustering of Word Embeddings}
    Top panels show clusters generated by the proposed dynamic word embedding method, trained on abstracts from 1994 to 2012 and 2022, respectively. 
    Word embeddings were obtained using a dynamic Word2Vec model trained on the respective set of abstracts.
    These embeddings were then reduced to two dimensions using UMAP, followed by clustering with a k-means algorithm. 
    The tables below each plot list the key concepts -- by proximity to the clusters center and frequency of occurrence in 2012 (2022) -- identified in each cluster.  
    Clusters generated from independently initialized dimensionality reduction schemes allow for analysis of concept relationships within the same year, however, cluster A0 in 2012 is not directly related to cluster B0 in 2022.
    Nonetheless, the results demonstrate that the learned word embeddings capture structured information about central topics in the field of quantum physics, illustrating how the landscape of research focus has evolved over the decade.}
    \label{fig:clustering}
\end{figure*}

\section{Methodology} \label{sec:method}
\subsection{Dataset}
The text corpus $\mathcal{D}$ consists of 66,839 abstracts from arXiv’s \emph{quant-ph} category, covering the period $ \Delta T $ from January 1994 to December 2023.
The vocabulary of all words in the corpus is denoted as $V$.  
We define the set of quantum physics concepts $C$ as multi-word tokens, each comprising one or more words $v \in V$ that represent distinct ideas or topics within the field.
To identify these concepts, we process the corpus by matching words $v$ to a predefined concept list as outlined in~\cite{guGenerationHumanexpertEvaluation2024}.
When a set of words matches a known quantum physics concept, it is replaced in the corpus with a single token, and the year of its appearance is recorded.
This approach results in a final list of 10,235 distinct quantum physics concepts and a modified text corpus $\tilde{\mathcal{D}}$ with an updated vocabulary $\tilde{V}$ containing the replaced tokens, where $C \subseteq \tilde{V}$.
Fig.~\ref{fig:overview}a illustrates how an abstract may contain specific quantum physics concepts, such as \emph{entanglement witnesses}, which are tracked as unique tokens.

\subsection{Dynamic Embedding Model} \label{sec:dynamic_model}
The goal of the dynamic embedding model is to obtain a vector $w_{c,t} \in \mathbb{R}^n $ for each token $ c \in C $ at each time step $t \in \Delta T$, encoding the semantic relationships between concepts in quantum physics literature and how these relationships change over time.
To achieve this, we use the Word2Vec model with the Skip-gram approach~\cite{mikolov2013efficient, mikolov2013distributed}.
Word2Vec is a neural network-based model that learns continuous vector representations of tokens (embeddings) from the text corpus $ \tilde{\mathcal{D}} $. 
The Skip-gram model is designed to predict the context tokens surrounding a given target word within a window size $ k $, learning embeddings that reflect semantic similarities based on context.
Conceptually, Skip-gram is a two-layer neural network trained to maximize the conditional probability $P(w_{i+j,t} | w_{i,t})$ of observing context token $w_{i+j,t}$ given a target word $w_{i,t}$~\cite{mikolov2013efficient}:

\begin{equation}
     \mathcal{L}_{\text{Skip}} = -\sum_{i \in \tilde{\mathcal{D}}} \sum_{\substack{-k \le j \le k \\ j \neq 0}} \log P(w_{i+j,t} | w_{i,t})
\end{equation}

The architecture consists of an input layer, a projection (hidden) layer, and an output layer.
To generate dynamically evolving embeddings, we train the model sequentially for each year in $ \Delta T $, initializing each step with the trained model from the previous year, as illustrated in Fig.~\ref{fig:overview}b.
This approach has proven to reveal semantic shifts over time~\cite{kimTemporalAnalysisLanguage2014}.
This process generates embeddings for the set of concepts $C$ from their first appearance in the text corpus up to the cutoff date. 
Since concepts emerge at different times in the corpus (see Fig.~\ref{appx:papers_per_year} inset), we use backtracking to maintain a consistent set of embeddings across all years. 
For concepts that first appear after the corpus’s starting year, we fill in the missing embeddings by using the embedding of their initial occurrence. 
This approach ensures that all concepts are trackable from the same initial year, providing a consistent basis for analysis over time.

\subsection{Predictive Model}
Given that the embedding features capture detailed information about relationships between various concepts and how these relationships evolve over time, we investigate training a machine learning model to predict whether concept pairs will eventually become connected based on their embedding data.
Our approach aims to predict if pairs of concepts, which have not yet appeared together in the same abstract, will co-occur in future research.
In our classification task, illustrated in Fig.~\ref{fig:overview}c, we categorize quantum physics concept pairs into two groups.
For both groups, the concept pairs have not co-occurred in the same abstract during a training window $ \Delta t $.
We assign the label $y=1$ if the concepts co-occur in the same abstract within a subsequent label window $\Lambda t$, and $ y=0 $ if they remain unconnected during this period.
To perform this classification, we use a neural network-based classifier.
The network is trained to differentiate between the two categories based on embeddings in the dataset $ \mathcal{X} $, minimizing the binary cross-entropy loss~\cite{RamosBCE2018} between true labels $ y $ and predicted labels $ \hat{y} $:
\begin{equation}
    \mathcal{L}_{\text{BCE}} = - \frac{1}{|\mathcal{X}|} \sum_{(x, y) \in \mathcal{X}} \left[ y \log(\hat{y}) + (1 - y) \log(1 - \hat{y}) \right] \label{eq:bce}
\end{equation}

\section{Results}\label{sec:result}
\subsection{Analyzing the Dynamic Embedding}
We begin by analyzing the quality of the proposed dynamic embedding method to assess its effectiveness in capturing relationships among quantum physics concepts.
To do this, we train the dynamic embedding model, as described in Sec.~\ref{sec:dynamic_model}, over two time frames: $\Delta t=[1994,2012]$ and $\Delta t=[1994,2022]$ with an embedding size of $n=128$.
To explore the distribution of the embeddings, we examine those from the final years, 2012 and 2022, each informed by data dating back to 1994.
We apply Uniform Manifold Approximation and Projection (UMAP)~\cite{mcinnes2018umap} to reduce the 128-dimensional embeddings to two dimensions, allowing us to observe relationships among quantum physics concepts in this reduced space. 
In this, UMAP aims to preserve a notion of both local and global structures from the original feature space~\cite{mcinnes2018umap}.
Additionally, we use k-means clustering~\cite{lloyd1982least} to identify nine distinct clusters within this reduced space, helping us analyze neighborhoods of related quantum physics concepts.
Both UMAP and k-means hyperparameters were optimized empirically for the clearest visualization.
The results of this clustering analysis are shown in Fig.~\ref{fig:clustering}, providing a visual representation of concept relationships and groupings based on their embeddings.
Below the figure, we include tables listing key concepts in each cluster, selected by proximity to the cluster center and their frequency in abstracts from 2012 (2022).

To evaluate embedding quality, we interpret the structural patterns within the clusters.
Each cluster contains thematically related quantum physics concepts; for instance, cluster $A0$ in 2012 focuses on quantum computing, while $A6$ contains concepts about stochastic noise.
Concept distances within clusters are also meaningful. For example, clusters $A5$ and $A8$, which are close, represent concepts related to experimental quantum physics, whereas $A5$ and $A7$, which are farther apart, include distinct topics.
This clustering shows that the embeddings capture structured information about central quantum physics topics.
To illustrate how research focus has shifted over time, we compare the clusters from 2012 to those from 2022. We observe emerging topics, such as gravitational waves and quantum advantage.
While these clusters allow us to interpret the structure of the embeddings within a single year, the independent initialization of dimensionality reduction and evolving embeddings over time mean that clusters from different years (e.g., $A0$ and $B0$) are not directly comparable.

The observed structure suggests that the embeddings encode meaningful information about relationships among quantum physics concepts and their evolution over time.

\subsection{Comparing the Performance of Embedding Methods}
The primary objective of this study is to evaluate the effectiveness of the proposed dynamic embedding technique for classifying quantum physics concept pairs and to compare its performance against existing methods. 
All results are presented in Tab.~\ref{table:comparison_results}, in which the performance of each model is evaluated using the area under the curve (AUC) metric~\cite{fawcett2004roc}.
The underlying receiver operating characteristic (ROC) curves are presented in Appx.~\ref{appx:comparison_results}.
To ensure an equal comparison, all embedding models -- whether utilizing our proposed embeddings or baseline methods -- were trained and evaluated using the same predictive (neural network) architecture, as outlined in Appx.~\ref{chap:training_details_classifier}. 
The only variable across experiments was the input representation, with the number of input neurons reflecting the dimensionality of the respective embeddings.

\begin{table*}[ht!] 
\centering
\begin{tabular}{>{\centering\arraybackslash}m{2.7cm}  >{\centering\arraybackslash}m{2cm}  >{\centering\arraybackslash}m{2cm} >{\centering\arraybackslash}m{3cm} >{\centering\arraybackslash}m{3.5cm} >{\centering\arraybackslash}m{3cm}}
\toprule[1.5pt] 
\textbf{Model} & \textbf{AUC} & \textbf{Reference} & \textbf{Knowledge Graph} & \textbf{Dynamic Embedding} & \textbf{Feature Type} \\
\midrule[1.5pt]
Word-Dynamic & \textbf{0.87} & Ours & No & Yes & Automatic \\
Word-Static & 0.79 &~\cite{mikolov2013distributed} & No & No & Automatic \\
Word-Hand & 0.57 &~\cite{tshitoyanUnsupervisedWordEmbeddings2019} & No & No & Handcrafted \\
Knowledge-Hand & 0.82 &~\cite{krennPredictingResearchTrends2020a} & Yes & No & Handcrafted \\
Knowledge-ProNE & 0.79 &~\cite{krennForecastingFutureArtificial2023a}M7A & Yes & No & Automatic \\
Knowledge-Node & 0.78 &~\cite{krennForecastingFutureArtificial2023a}M7B & Yes & No & Automatic \\
\bottomrule[1.5pt] 
\end{tabular}

\caption{\textbf{Comparing Performance of Embedding Techniques:} 
Area under the curve (AUC) values for six embedding methods, comparing word embeddings and traditional knowledge graph approaches.
Abbreviations are: the proposed dynamic word embeddings (Word-Dynamic), 
static word embeddings (Word-Static), 
static word embeddings with hand crafted features (Word-Hand),
Knowledge Graph with hand-crafted features (Knowledge-Hand), 
Knowledge Graph with machine-learned (ProNE) features (Knowledge-Prone), 
and Knowledge Graph with machine-learned (Node2Vec) features (Knowledge-Node), 
Models are trained to predict concept co-occurrence in the period $\Lambda t=[2018,2020]$ based on embeddings from $\Delta t=[1994,2017]$, the displayed performance is evaluated with a 3-year time shift.
}
\label{table:comparison_results}
\end{table*}

We trained and validated the models using embeddings generated from abstracts spanning the years $\Delta t=[1994,2017]$, with label windows corresponding to the years $\Lambda t=[2018,2020]$. 
For testing, we utilized embeddings up to $\Delta t=[1994,2020]$, validating predictions against the label window covering $\Lambda t=[2021,2023]$.

The results demonstrate the proposed dynamic embedding method consistently surpassing the baseline approaches. 
Since both features were derived from the same underlying dataset and the predictive models were trained with identical structures, the performance disparity can be attributed solely to the input embeddings. 

We conclude that the model's capacity to utilize the temporal structure of these embeddings enriches the contextual information, leading to more accurate and robust predictions. 
In the following, we briefly outline the distinctions between each embedding technique, with a more detailed description provided in Appx.~\ref{appx:embedding}.

\noindent \textbf{(Ours) Dynamic Word Embedding (Word-Dynamic)} --
Our proposed dynamic embedding approach represents each concept as a vector $w_{c,t} \in \mathbb{R}^n $ for each token $ c \in C $ at each time step $t \in \Delta T$ with $n=128$.
It captures the semantic relationships between concepts in quantum physics literature and how these relationships evolve over time. 
While this technique does not explicitly encode the connections between concepts, it provides the predictive model with comprehensive information about the structure within the embedding space.

\noindent \textbf{Static Word Embedding (Word-Static)} --
The static word embedding approach can be viewed as the standard Word2Vec method for generating embeddings $w_{c} \in \mathbb{R}^n $ with $n=128$.
Like the dynamic embedding, the training is performed in an unsupervised manner; however, it uses a single corpus of abstracts from a specified time frame, $\Delta t$. 
Thus, it does not incorporate the dynamic training introduced in Sec.~\ref{sec:dynamic_model}.
The resulting structure encodes the relationships between concepts implicitly within an embedding space, but does not account for temporal changes in these relationships.

\noindent \textbf{Static Word Embedding and Handcrafted Feature (Word-Hand)} --
This method utilizes the same static word embeddings but derives a handcrafted feature: the cosine similarity between word embedding vectors. In~\cite{tshitoyanUnsupervisedWordEmbeddings2019}, this metric is used to predict potential undiscovered connections in material science, effectively identifying novel relationships by assessing the proximity of concepts in the embedding space. 
While this method also implicitly encodes relationships between concepts, it simplifies the information into a single handcrafted feature.

\noindent \textbf{Knowledge Graph and Handcrafted Feature (Knowledge-Hand)} --
This method does not use word embeddings but instead creates an evolving knowledge graph from the abstracts. 
In this graph, concepts are represented as nodes, and connections between concepts—i.e., if they co-occur in the same abstract—are modeled as edges. 
In~\cite{krennPredictingResearchTrends2020a}, 15 handcrafted features are extracted from the knowledge graph to predict the connection between concepts. 
The approach formulates the task as a link prediction problem, training a machine learning model to classify whether two nodes, representing distinct concepts, will form a link within a specified timeframe. 
In general, the knowledge graph representation abstracts away all information aside from the connections, focusing explicitly on modeling the relationships between concepts.

\noindent \textbf{Knowledge Graph and Machine Learning Feature (Knowledge-Node)} --
Instead of selecting handcrafted features from knowledge graphs for the classification task, machine learning-based features can also be used~\cite{krennForecastingFutureArtificial2023a}. 
This model is based on Node2Vec embeddings~\cite{grover2016node2vecscalablefeaturelearning}, which eliminates the need for handcrafted features in link prediction within knowledge graphs.

\noindent \textbf{Knowledge Graph and Machine Learning Feature (Knowledge-ProNE)} -- 
This model also uses machine-learned embeddings from the knowledge graph, in this case the ProNE method~\cite{zhang2019prone}.
Thus, it also incorporates these automated embeddings into a neural network framework for link prediction between concepts, without the need for explicit feature engineering.

\subsection{Analyzing the Model Calibration} 
\begin{figure}[!t]
    \centering
    \includegraphics[width=0.5\textwidth]{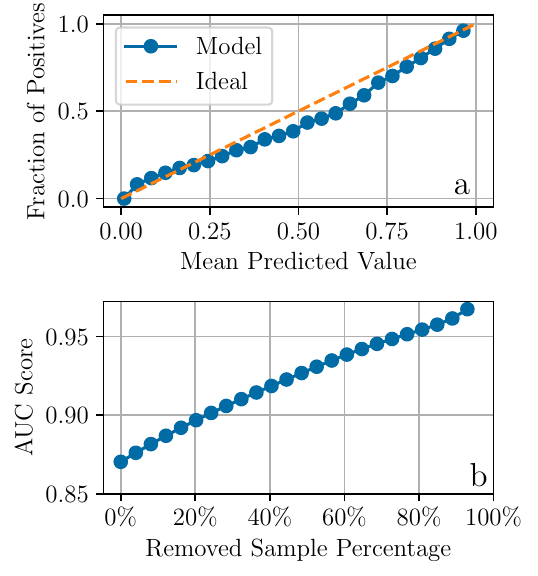} 
    \caption{\textbf{Model Confidence:} 
    Predictive model trained on the proposed embedding for $\Delta t=[1994,2017]$, tested to predict data in $\Lambda t=[2020,2023]$. 
    (a) Calibration plot showing the probability of the model's predictions being correct, with a comparison to a perfectly calibrated model (orange dashed line). 
    The model is well-calibrated for predictions near probabilities of 0 and 1, confidently classifying these samples. 
    (b) AUC score as a function of the fraction of low-confidence predictions discarded. The plot illustrates that removing uncertain samples enhances the AUC score.
    }
    \label{fig:callibration_performance}
\end{figure}

In the previous section, we trained the predictive model with a logistic output by minimizing cross-entropy loss with binary labels, as introduced in Eq.~\ref{eq:bce}, allowing the output to be interpreted as the probability that two quantum physics concepts will be linked in the future. 
This probability is generally well-calibrated but shows increased uncertainty around predicted probabilities near 0.5, as illustrated in Fig.~\ref{fig:callibration_performance}a. 
For example, among samples with an output value of 0.9, approximately 90\% have a true label of 1. 
Predictions closer to 0 or 1 indicate higher confidence, while probabilities near 0.5 signal greater uncertainty and an increased likelihood of misclassification. 
Therefore, this probabilistic output thus enables selection of test samples, allowing uncertain samples to be filtered out.
In Fig.~\ref{fig:callibration_performance}b, we see that removing low-confidence predictions, such as the 20\% most uncertain pairs, increases the AUC score beyond 0.9 - a threshold indicating strong predictive performance~\cite{hosmer2013applied}.
For studying interesting concept combinations, discarding a substantial portion of samples is not inherently problematic. 
Focusing on a small subset of high-confidence predictions makes investigating samples that warrant further analysis more feasible. 
This approach makes confidence-based selection an effective strategy for identifying this promising subset.

\subsection{Analyzing the Model Predictions} 
An added benefit of dynamic word embeddings paired with a well-calibrated predictive model is their ability to capture and interpret the evolution of research concepts over time. 
Thus, as a final step, we evaluate whether our model -- trained on quantum physics abstracts from prior years -- could predict concept combinations that would later emerge in the literature.

\begin{figure}[t!]
    \centering
    \includegraphics[width=\linewidth]{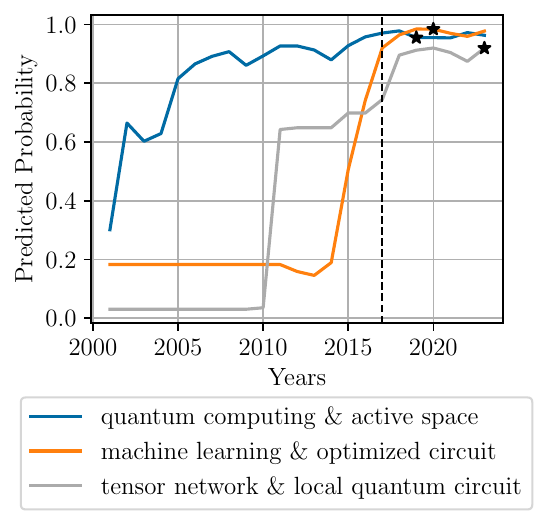}
    \caption{\textbf{Validating Past Prediction:}
    Evolution of the prediction probability for three distinct quantum physics concept pairs in a model trained on embeddings up to 2017. 
    Markers indicate the year of the first published abstract containing these concept pairs, as referenced in Refs.~\cite{Bauman_2019,Cincio_2021,liu2024simulating2dtopologicalquantum}.}
    \label{fig:comparison}
\end{figure}

In Fig.~\ref{fig:comparison}, we visualize the predicted probability -- shown to be well-calibrated for confident cases in the previous section -- of a model trained on dynamic embeddings from $\Delta t=[1994,2017]$ and then evaluated on the embeddings for each year through 2023. 
We illustrate three pairs of concepts, showing how the model’s predictions evolved over time.  
The star marker in each line indicates the year when the concept pair first appeared together in a published abstract.
The selected pairs highlight recent popular combinations in the field:
(1) The predicted link between \emph{quantum computing} and \emph{active space} methods first appears in Ref.~\cite{Bauman_2019} given our text corpus and concept list.
(2) The predicted connection of using \emph{machine learning} for constructing \emph{optimized circuits} for noise-resilience first appears in Ref.~\cite{Cincio_2021}.
(3) The technique of using \emph{tensor network} approaches in investigating \emph{local quantum circuits} first appears in Ref.~\cite{liu2024simulating2dtopologicalquantum}.
Importantly, the concept set $C$ is highly granular and does not yet account for synonyms; we discuss the implications of this limitation in Sec.~\ref{sec:discussion}.
Nonetheless, this visualization shows that the model's increasing probability predictions align well with the first recorded uses of these concept pairs in the corpus.
While the pairs were chosen based on rising topics of interest, we note that high predicted probabilities also corresponded to these papers having (1) 88, (2) 118, and (3) 9 citations, respectively, as of this manuscript’s writing.

\section{Discussion and Outlook}\label{sec:discussion}
In this work, we demonstrated the application of dynamic word embeddings for predicting research trends using neural networks, with a focus on quantum physics concepts. 
Our results show that features generated from an unsupervised embedding model -- without the need for direct human input -- show promising performance, leading to improved prediction accuracy compared to the baseline embeddings.
Thus, the presented embedding method offers an interesting research direction for leveraging machine learning to investigate how experts across diverse fields approach or solve structurally similar problems.

\noindent \textbf{Benchmark Additional Dynamic Embedding Approaches} --
For this initial study, we adopted the dynamic embedding method from Ref.~\cite{kimTemporalAnalysisLanguage2014}. 
A natural extension of this work would be to compare the effectiveness of various other dynamic embedding techniques for the specific prediction task at hand~\cite{bamlerDynamicWordEmbeddings2017,hornExploringWordUsage2021}.  

\noindent \textbf{Benchmark LLM Embeddings} --
Additionally, exploring the potential of more modern embedding models like LLaMA~\cite{touvron2023llama}, BERT~\cite{devlinBERTPretrainingDeep2019a} or GPT~\cite{brownLanguageModelsAre2020} could further enhance the performance of our method.
We opted not to incorporate these advanced models in the present study for the following reasons. 
Modern language models generate contextual embeddings, which vary based on the surrounding text in which a word appears. 
Unlike traditional models like Word2Vec, which produce a static embedding for each word, contextual embeddings can capture subtle changes in meaning depending on the context. 
While this greatly enhances the representational power of such models, applying them to our scenario would introduce additional complexity.
In quantum physics research, where concepts can be used in different contexts, contextual embeddings would result in multiple representations of the same concept. 
This would require further processing, such as averaging the representations to generate a consistent embedding for each concept, adding a layer of complexity to the analysis. 
We leave this refinement for future research.

\noindent \textbf{Extended Model Selection} --
A valuable future research direction involves comprehensive model selection and optimization, customizing the neural network architecture to suit the specific characteristics of the dynamic embedding features.
This approach could further improve predictive performance and underscore the potential of dynamic embeddings.
We focused on using the embedding from the final year of the training window, as empirical investigations showed that incorporating multiple time slices did not significantly impact the final AUC score. 
A future study could explore whether a deep recurrent neural network architecture can leverage information from multiple time slices for enhanced predictive performance.

\noindent \textbf{Application for Quantum Physics Research} --
The ultimate goal of the proposed method is to forecast the future trajectory of scientific research by predicting potential directions that scientists might explore. 
Our method represents a step towards achieving high-quality predictions in this domain. 
In Fig.~\ref{fig:comparison}, we demonstrate that our method assigns high probabilities to fruitful concept combinations.

The predictions depend on the specific concepts extracted from each abstract, especially for very prominent concepts that have many synonyms (different phrases with the same semantic meaning). 
An interesting next step would  hierarchical grouping of related concepts (synonyms), meaning that although terms like \emph{active space} and \emph{quantum algorithm} may have co-occured before, the model only flagged the exact combination of \emph{quantum computing} and \emph{active space} in the displayed year. 
This would reduce the granularity of the set of concepts but could enhance the model’s ability to track when distinct concept areas begin to intersect.

\section*{Data \& Code availability}
The code for this study, including preprocessing, training, visualization details, and concept list are available in the associated \href{https://github.com/FelixFrohnertDB/arxiv_nlp}{GitHub repository}. 
The arXiv abstract corpus can be accessed at \href{https://www.kaggle.com/datasets/Cornell-University/arxiv}{Kaggle}.

\section*{Acknowledgments}
The authors would like to thank Simon C. Marshall, Stefano Polla, and Patrick Emonts for useful discussions.
This work was supported by the Dutch National Growth Fund (NGF), as part of the Quantum Delta NL program.

\appendix

\setcounter{figure}{0}
\renewcommand{\thefigure}{A\arabic{figure}}

\begin{figure}[b!]
    \centering
    \includegraphics[width=\linewidth]{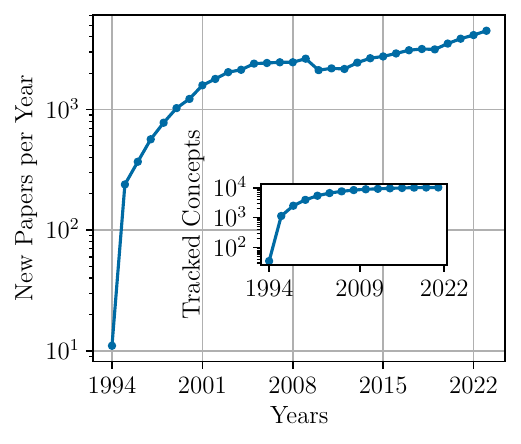}
    \caption{\textbf{The number of Quantum-Physics Pre-Prints per year is growing:}  The number of (main) papers publishes on the ArXiv, as well as (inset) the key concepts within them is steadily increasing each year. 
    }
    \label{appx:papers_per_year}
\end{figure}

\begin{figure}[t!]
    \centering
    \includegraphics[width=\linewidth]{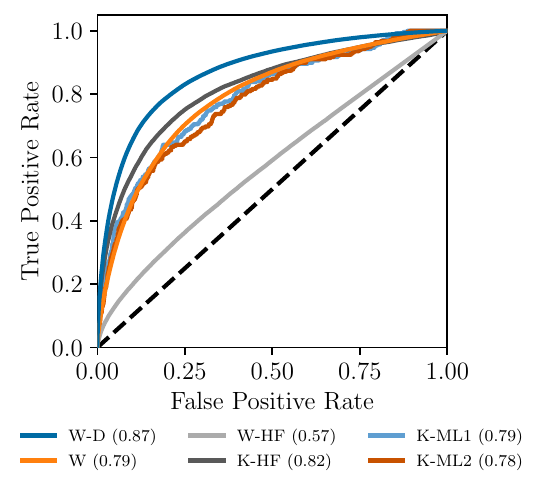}
    \caption{\textbf{Comparing Performance of Embedding Techniques:} 
This figure presents the area under the ROC curve values for six embedding methods, comparing word embeddings and traditional knowledge graph approaches.
Abbreviations are: the proposed dynamic word embeddings (W-D), 
static word embeddings (W), 
static word embeddings with hand crafted features (W-HF),
Knowledge Graph with hand-crafted features (K-HF), 
Knowledge Graph with machine-learned (ProNE) features (K-ML1), 
and Knowledge Graph with machine-learned (Node2Vec) features (K-ML2), 
Models are trained to predict concept co-occurrence in the period $\Lambda t=[2018,2020]$ based on embeddings from $\Delta t=[1994,2017]$, the displayed performance is evaluated with a 3-year time shift.
}
    \label{appx:comparison_results}
\end{figure}
\section{Details on Embedding Methods} \label{appx:embedding}

\subsection{Static Word Embeddings}
For both (Word-Static) and (Word-Hand) methods, we utilized a Word2Vec model configured in the same way as described in Appx.~\ref{chap:training_details_embedding}. 
The primary distinction between the proposed dynamic embedding model and the static embedding approach lies in their training processes.
In the case of static embeddings, the training is conducted in a single run using the complete text corpus over the entire time period $\Delta t$. 
This approach captures the overall context and relationships within the data, producing a fixed set of word vectors that do not change over time.
Conversely, the dynamic embedding model is designed to be updated iteratively. 
It is trained and initialized anew after each year within the specified interval $\Delta t$. 
This allows the model to adapt to shifts in language usage and context that may occur over time, resulting in a more nuanced representation of word meanings that reflects temporal changes in the corpus.

\subsection{Knowledge Graph and Handcrafted Feature}
The following embedding methods we compare our proposed approach to are fundamentally based on knowledge graphs. 
The concept of representing combinations of concepts as semantic networks was first introduced in Ref.~\cite{rzhetsky2015choosing} and later applied to a prediction in task in Ref.~\cite{krennPredictingResearchTrends2020a}.
For benchmarking, we construct the knowledge graph as follows: whenever two concepts co-occur in an abstract within the timeframe $\Delta t$, we create a link between the corresponding nodes, representing an edge in the graph. The weight of each edge is determined by the number of connections between the nodes.

For the classification task, the model presented in~\cite{krennPredictingResearchTrends2020a} relies on $15$ handcrafted features derived from pairs of nodes $v_1$ and $v_2$ in the graph. 
These features include:
The degrees of $v_1$ and $v_2$ in the current year and the previous two years ($6$ features).
The number of shared neighbors of $v_1$ and $v_2$ in the current year and the previous two years ($6$ features).
The number of shared neighbors between $v_1$ and $v_2$ in the current year and the previous two years ($3$ features).

\subsection{Knowledge Graph and Machine Learning Feature}
In general, selecting handcrafted features preselects information which might not be ideal for a downstream machine learning task.
Thus, we include Models M7A and M7AB from Ref.~\cite{krennForecastingFutureArtificial2023a}, which perform link prediction while eliminating the need for handcrafted features. Model M7A utilizes Node2Vec embeddings~\cite{grover2016node2vecscalablefeaturelearning}, a technique that learns low-dimensional representations of nodes in a graph based on their network structure, facilitating the capture of contextual relationships. 
Model M7B employs ProNE embeddings~\cite{zhang2019prone}, which extend Node2Vec by incorporating proximity information to enhance the quality of the embeddings. 
Both models operate within a neural network framework, leveraging these automated embeddings to predict links between concepts without the explicit feature engineering that previous methods required.

\section{Embedding Training Details} \label{chap:training_details_embedding}
The embedding model was implemented using the Gensim Python library~\cite{rehurek2011gensim}. 
The context window size, which defines the number of context tokens to the left and right of the target word, was set to $k=10$.
The dimensionality of the word embeddings is a crucial hyperparameter that influences the model's ability to capture nuanced semantic relationships. 
After experimenting with various dimensions, we selected a vector size of $n=128$ dimensions, providing balance between embedding quality and computational efficiency for the downstream prediction task.
We utilized a corpus of abstracts from arXiv quantum physics papers, which were preprocessed to extract relevant concepts. 
The corpus was tokenized and lemmatized, with concepts treated as individual tokens.

\section{Classifier Training Details} \label{chap:training_details_classifier}
The predictive model employed a neural network designed to forecast future connections between quantum physics concepts and was implemented in PyTorch~\cite{paszkePyTorchImperativeStyle2019}. 
The model is a multi-layer perceptron (MLP) consisting of a series of three layers.
The network uses Parametric ReLU activations, batch normalization, and dropout to mitigate overfitting. 
The final output layer applies ReLU and sigmoid activations to produce a binary classification probability.
The training process incorporates early stopping to avoid overfitting. 
Additionally, a learning rate scheduler is employed to dynamically adjust the learning rate during training, enhancing convergence and model performance.

To address class imbalance, where the number of negative (non co-occurring) concepts substantially exceeds the number of positive (occurring) concepts, we applied simple uniform sampling on the negative examples to balance the class distribution. 
This approach ensured that our model was trained on a more representative dataset and improved the reliability of our evaluation metrics.

\newpage
\bibliography{reference.bib}

\end{document}